\documentclass{article}       
\usepackage{authblk}
\usepackage{amsmath}
\usepackage{amsfonts}
\usepackage{dsfont}  
\usepackage{pdfpages}
\usepackage{pgf}
\usepackage[letterpaper]{geometry}
\usepackage{natbib}
\usepackage{graphicx}
\usepackage{multirow}
\usepackage{amssymb}
\usepackage{amsthm}
\usepackage{amscd}
\usepackage{times}
\usepackage{booktabs}
\usepackage{url}
\usepackage{graphicx}
\usepackage{framed,multirow}
\usepackage{bm}
\usepackage{latexsym}

\usepackage[doublespacing]{setspace}

\def\x{{\bf x}}
\def\X{{\bf X}}

\newcommand{\indicator}[1]{\mathds{1}_{\left[ {#1} \right] }}

\title{Deep Mixtures of Unigrams for uncovering Topics in Textual Data}

\author[1]{Cinzia Viroli\footnote{mailto: \texttt{cinzia.viroli@unibo.it}}} 
 \author[1]{Laura Anderlucci}

\affil[1]{Department of Statistical Sciences, University of Bologna, Italy.}
\date{}

\begin{document}

\maketitle

\begin{abstract}
Mixtures of Unigrams 
are one of the simplest and most efficient tools for clustering textual data, as they assume that documents related to the same topic have similar distributions of terms, naturally described by Multinomials. When the classification task is particularly challenging, such as when the document-term matrix is high-dimensional and extremely sparse, a more composite representation can provide better insight on the grouping  structure. In this work, we developed a deep version of mixtures of Unigrams for the unsupervised classification of very short documents with a large number of terms, by allowing for models with further deeper latent layers; the proposal is derived in a Bayesian framework. The behaviour of the Deep Mixtures of Unigrams is empirically compared with that of other traditional and state-of-the-art methods, namely $k$-means with cosine distance, $k$-means with Euclidean distance on data transformed according to Semantic Analysis, Partition Around Medoids, Mixture of Gaussians on semantic-based transformed data, hierarchical clustering according to Ward's method with cosine dissimilarity, Latent Dirichlet Allocation, Mixtures of Unigrams estimated via the EM algorithm, Spectral Clustering and Affinity Propagation clustering. The performance is evaluated in terms of both correct classification rate and Adjusted Rand Index. Simulation studies and real data analysis prove that going deep in clustering such data highly improves the classification accuracy.
\end{abstract}


\section{Introduction}\label{sec0}

Deep learning methods are receiving an exponentially increasing interest in the last years as powerful tools to learn complex representations of data. They can be basically defined as a multi-layer stack of algorithms or modules able to gradually learn a huge number of parameters in an architecture composed by multiple nonlinear transformations \citep{lecun}. Typically, and for historical reasons, a structure for deep learning is identified with advanced neural networks: deep Feed Forward, Recurrent, Auto-encoder, Convolution neural networks are very effective and used algorithms of deep learning \citep{schmidhuber}.
They demonstrated to be particularly successful in supervised classification problems arising in several fields such as image and speech recognition, gene expression data, topic classification.
Under the framework of graph-based learning, \cite{peng2016constructing} proposed an efficient method to produce robust subspace clustering and subspace learning; deep model-Structured AutoEncoder for subspace clustering were introduced by \cite{peng2018structured} to map input data points into nonlinear latent
spaces while preserving the local and global subspace structure. \cite{zhou2018transfer} addressed the data sparsity issue in hashing.

When the aim is uncovering unknown classes in an unsupervised classification perspective, important methods of deep learning have been developed along the lines of mixture modeling, because of their ability to decompose a heterogeneous collection of units into a finite number of sub-groups with homogeneous structures \citep{Fraley,Peel}.  In this direction, \cite{NIPS2014} proposed Multilayer Gaussian Mixture Models for modeling natural images; \cite{Tang12} defined deep mixture of factor analyzers with a greedy layer-wise learning algorithm able to learn each layer at a time. \cite{Viroli2017} developed a general framework for Deep Gaussian mixture models that generalizes and encompasses the previous strategies and several flexible model-based clustering methods such as mixtures of mixture models \citep{Li2005}, mixtures of Factor Analyzers \citep{MCLACHLAN2003}, mixtures of factor analyzers with common factor loadings \citep{Baek}, heteroscedastic factor mixture analysis \citep{Montanari2010} and mixtures of factor mixture analyzers introduced by \cite{Viroli2010}. A general `take-home-message' coming from the existing deep clustering strategies is that deep methods vs shallow ones appear to be very efficient and powerful tools especially for complex high-dimensional data; on the contrary, for simple and small data structures, a deep learning strategy cannot improve performance of simpler and conventional methods or, to better say, it is like to use a `sledgehammer to crack a nut' \citep{Viroli2017}.

The motivating problem behind this work derives from ticket data (i.e. content of calls made to the customer service) of an important mobile phone company, collected in Italian language. When a customer calls the assistance service, a ticket is created: the agent classifies it as e.g. a claim, a request of information for specific services, deals or promotions. Our dataset consists of tickets related to five classes of services, previously classified from independent operators. The aim is to define an efficient clustering strategy to automatically assign the tickets into the same classes without the human judgment of an operator. The data are textual and information are collected in a document-term matrix with raw frequencies at each cell. They have a very complex and a high-dimensional structure, caused by the huge number of tickets and terms used by people that call the company for a specific request and by a relevant degree of sparsity (after a pre-processing step, the tickets have indeed an average length of only 5 words and, thus, the document-term matrix contains zero almost everywhere).

The simplest topic model for clustering document-term data is represented by Mixture of Unigrams \citep[MoU,][]{Nigam}, based on the idea of modeling the word frequencies as multinomial distributions, under the assumption that a document refers to a single topic. Table \ref{tab.acc2} shows that on the ticket dataset this method appears to be the most efficient tool for classifying the complex ticket data, compared to other conventional clustering strategies such as $k$-means, Partition Around Medoids and hierarchical clustering. The reason is probably related to the fact that, by using proportions, MoU is not affected by the large amount of zeros, differently from the other competitors. We also compared MoU with the Latent Dirichlet Allocation model \citep[LDA,][]{blei2003latent}, which represents an important and very popular model in textual data analysis, allowing documents to exhibit multiple topics with a different degree of importance. The Latent Dirichlet Allocation model has demonstrated great success on long texts \citep{Griffiths5228}, and it could be thought of as a generalization of the MoU, because it adds a hierarchical level to it and, hence, much more flexibility. However, when dealing with very short documents like the ticket dataset, it is very rare that a single unit could refer to more than one topic; in such cases, the LDA model may not improve the clustering performance of MoU.

The aim of this paper is to derive a deep generalization of Mixtures of Unigrams, in order to better uncover topics or groups in case of complex and high-dimensional data. The proposal will be derived in a Bayesian framework and we will show that it will be particularly efficient for classifying the ticket data with respect to the `shallow' MoU model. We will also show the good performance of the proposed method in a simulation study.

The paper is organized as follows. In the next section the mixture of unigrams model is described. In Section 3 the deep formulation of the model is developed. Section 4 is devoted to the estimation algorithm for fitting the model. Experimental results on simulated and real data are presented in Sections 5 and 6, respectively. We conclude the paper with some final remarks (Section 7).

\section{Mixtures of Unigrams}\label{sec1}
Let $\X$ be a document-term matrix of dimension $n \times T$ containing the word frequencies of each document in row and let $k$ be the number of homogeneous groups in which documents could be allocated. Let $\x_d$ be the single document of length $T$, with $d=1,\ldots,n$.

In MoU the distribution of each document has a specific distribution function conditional on the values of a discrete latent allocation variable $z_d$ describing the probability of each topic. More precisely,

\begin{eqnarray}\label{mou}
  p(\x_d)=\sum_{i=1}^k \pi_i p(\x_d|z_d=i),
\end{eqnarray}

\noindent with $p(z_d=i)=\pi_i$, under the restrictions (i) $\pi_i>0$ for $i=1,\ldots,k$ and (ii) $\sum_{i=1}^k \pi_i=1$.

The natural distribution for $p(\x_{d}|z_d=i)$ is represented by the multinomial distribution with a parameter vector, say $\boldsymbol\omega_i$, that is cluster-specific:
\begin{eqnarray*}
  p(\x_{d}|z_d=i)= \frac{N_d!}{\prod_{t=1}^{T} x_{dt}!} \prod_{t=1}^T \omega_{ti}^{x_{dt}},
\end{eqnarray*}

\noindent with $N_d=\sum_{t=1}^{T} x_{dt}$ denoting the word-length of the document $d$. The multinomial distribution assumes that, conditionally to the cluster membership, all the terms can be regarded as independently distributed.

The model is indeed a simple mixture of multinomial distributions that can be easily estimated by the EM algorithm \citep[see][for further details]{Nigam} under the assumption that a document belongs to a single topic and the number of groups coincides with the number of topics. The approach has been successfully applied not only to textual data,  where it originated from, but to genomic data analysis. In this latter field a particular improvement of the method consists in relaxing the conditional independence assumption of the variables/terms, by using $m$-order Markov properties, thus leading to the so-called \emph{m-gram} models \citep{tomovic2006n}.
Due to the limited average length of documents in ticket data, co-occurrence information is very rare and this extension would be not effective on these data.

\section{Going deep into mixtures of unigrams: a novel approach}

We aim at extending MoU by allowing a further layer in the probabilistic generative model, so as to get a nested architecture of nonlinear transformations able to describe the data structure in a very flexible way. At the deepest latent layer, the documents can come from $k_2$ groups with different probabilities, say $\pi_j^{(2)} \ \ j=1,\ldots,k_2$. Conditionally to what happened at this level, at the top observable layer the documents can belong to $k_1$ groups with conditional probabilities
$\pi_{i|j}^{(1)} \ \ i=1,\ldots,k_1$. For the sake of a simple notation, we refer in the following to a generic document denoted by $\x$.
The distribution of $\x$ conditionally to the two layers is a multinomial distribution with cluster-specific proportions $\boldsymbol\omega$:

\begin{eqnarray}\label{DMoU.obs}
  p(\x|z^{(1)}=i,z^{(2)}=j,\boldsymbol\omega)= \frac{\left(\sum_{t=1}^{T} x_{t}\right)!}{\prod_{t=1}^{T} x_{t}!} \prod_{t=1}^T \omega_{tij}^{x_{t}},
\end{eqnarray}

\noindent where $z^{(1)}$ and $z^{(2)}$ are the allocation variables at the top and at the bottom layers, respectively. They are discrete latent variables that follow the distributions:

\begin{eqnarray}\label{DMoU.pesi1}
  p(z^{(2)}=j) = \pi_j^{(2)},
\end{eqnarray}
and

\begin{eqnarray}\label{DMoU.pesi2}
  p(z^{(1)}=i|z^{(2)}=j)=\pi_{i|j}^{(1)}.
\end{eqnarray}

In mixture of unigrams the proportions $\boldsymbol\omega$ were fixed parameters. Here we assume they are realizations of random variables with a Dirichlet distribution. In order to have a flexible and deep model we assume that the parameters of the Dirichlet distribution are a linear function of two sets of parameters that originate in the two subsequent layers. The Dirichlet parameters are $\boldsymbol\beta_i + \boldsymbol\alpha_j\boldsymbol\beta_i=\boldsymbol\beta_i(1+\boldsymbol\alpha_j)$, where $\boldsymbol\beta_i$ and  $\boldsymbol\alpha_j$ are vectors of length $T$.
Since they must be positive and the overall model must be identifiable, we further assume that $\beta_{it}>0$ and $-1 < \alpha_{jt} < 1$. These choices lead to a nice interpretative perspective: at the bottom layer $\boldsymbol\alpha_j$ acts as a perturbation on the cluster-specific $\boldsymbol\beta_i$ parameters of the top layer.
Therefore the distribution is:

{\small
\begin{eqnarray}\label{DMoU.lat}
  p(\boldsymbol\omega|z^{(1)}=i,z^{(2)}=j) =\frac{\Gamma\left(\sum_{t=1}^T \beta_{it}(1+\alpha_{jt})\right)}{\prod_{t=1}^T \Gamma(\beta_{it}(1+\alpha_{jt}))} \prod_{t=1}^T \omega_{tij}^{\beta_{it}(1+\alpha_{jt})-1},
\end{eqnarray}
}

\noindent where $\Gamma$ denotes the Gamma function.

\begin{figure}[t]
\centerline{\includegraphics[width=0.9\textwidth]{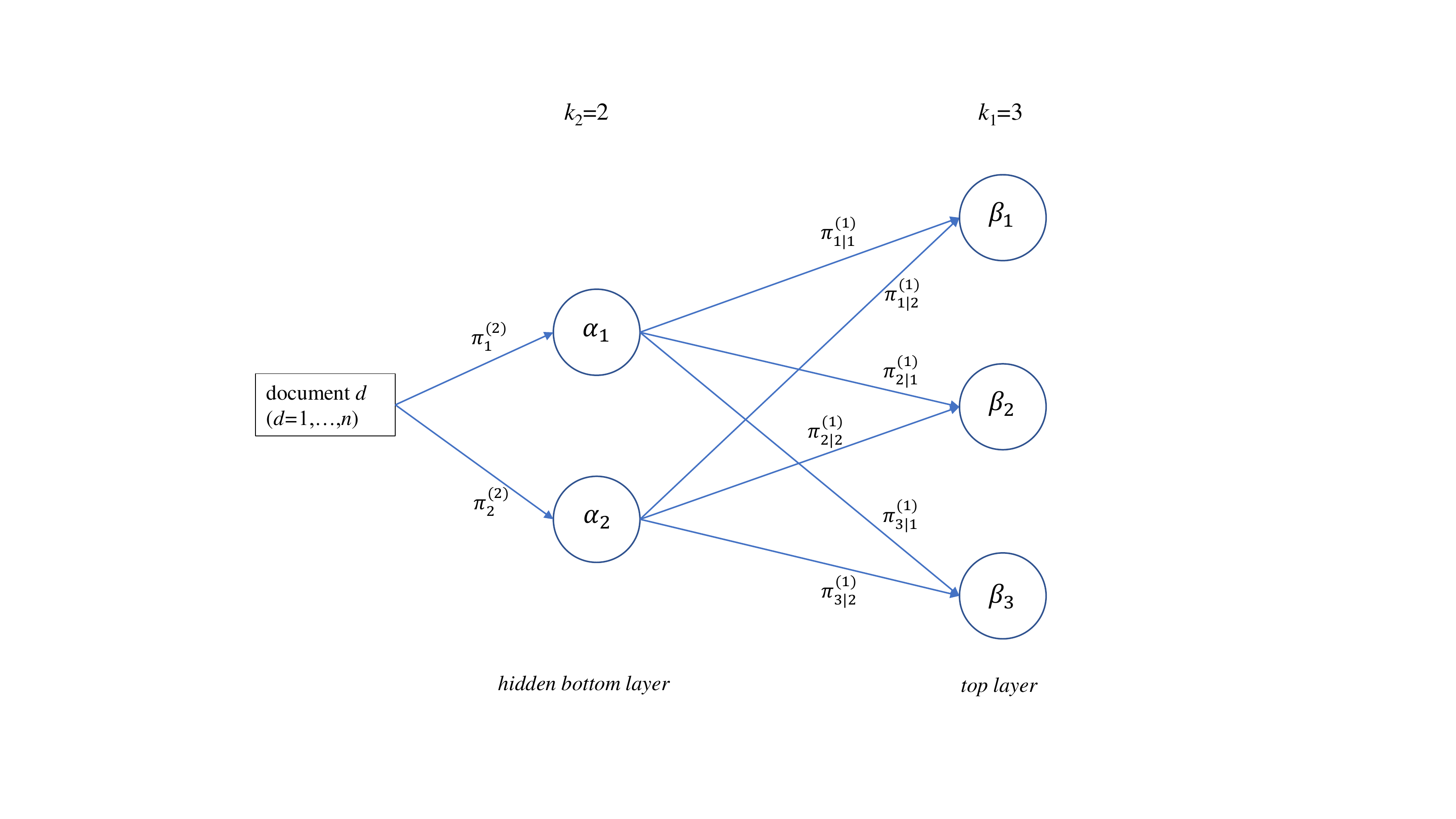}}
\caption{Structure of a Deep MoU with components $k_1=3$ and $k_2=2$. \label{fig1}}
\end{figure}

An example of Deep MoU structure is depicted in Figure \ref{fig1} for the case $k_1=3$ and $k_2=2$. Notice that in a model with $k_1=3$ and $k_2=2$ components we have an overall number $M$ of sub-components equal to 5, but $6>M$ possible paths for each document. The paths share and combine the parameters of the two levels, thus achieving great flexibility with less parameters to be estimated.

By combining equations (\ref{DMoU.obs}) and (\ref{DMoU.lat}), the latent variable $\boldsymbol\omega$ can be integrated out from the model estimation, thus gaining efficiency without losing flexibility and interpretability. More precisely:

{\small
\begin{eqnarray}\label{DMoU.out}
\begin{aligned}
p&(\x|z^{(1)}=i,z^{(2)}=j) \\
  &= \int p(\x|z^{(1)}=i,z^{(2)}=j,\boldsymbol\omega)p(\boldsymbol\omega|z^{(1)}=i,z^{(2)}=j) d \boldsymbol\omega \\
  &= \frac{\left(\sum_{t=1}^{T} x_{t}\right)!}{\prod_{t=1}^{T} x_{t}!}\frac{\Gamma\left(\sum_{t=1}^T \beta_{it}(1+\alpha_{jt})\right)}{\prod_{t=1}^T \Gamma(\beta_{it}(1+\alpha_{jt}))}\int  \prod_{t=1}^T \omega_{tij}^{x_{t}+\beta_{it}(1+\alpha_{jt})-1} d \boldsymbol\omega_{ij} \\
  &= \frac{\left(\sum_{t=1}^{T} x_{t}\right)!}{\prod_{t=1}^{T} x_{t}!}\frac{\Gamma\left(\sum_{t=1}^T \beta_{it}(1+\alpha_{jt})\right)}{\prod_{t=1}^T \Gamma(\beta_{it}(1+\alpha_{jt}))} \frac{\prod_{t=1}^T x_{t}+\Gamma(\beta_{it}(1+\alpha_{jt}))}{\Gamma\left(\sum_{t=1}^T x_{t}+\beta_{it}(1+\alpha_{jt})\right)}\\
  &= \frac{\sum_{t=1}^{T} x_{t}}{\prod_{t=1}^{T} x_{t}} \frac{B\left(\sum_{t=1}^T x_t,\sum_{t=1}^T \beta_{it}(1+\alpha_{jt})\right)}{\prod_{t=1}^T B\left( x_t, \beta_{it}(1+\alpha_{jt})\right)},
  \end{aligned}
\end{eqnarray}
}
where $B$ denotes the Beta function.

\begin{figure}[t]
\centerline{\includegraphics[width=0.9\textwidth]{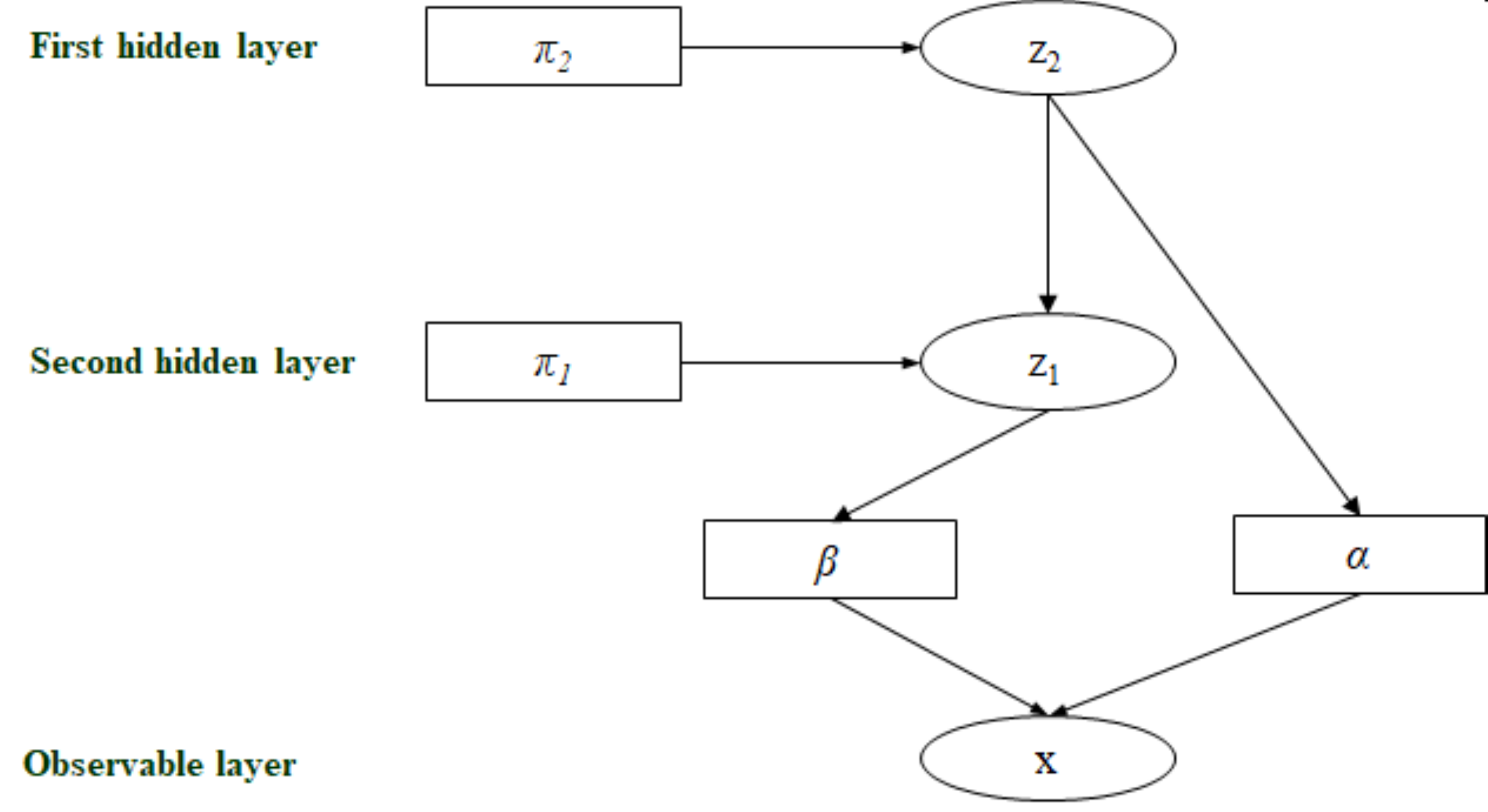}}
\caption{DAG specifying the Deep MoU model. \label{fig2}}
\end{figure}

Figure \ref{fig2} shows the Directed Acyclic Graph (DAG) summarizing the dependence
structure of the model.

Finally by combining formulas (\ref{DMoU.out}) with (\ref{DMoU.pesi1}) and (\ref{DMoU.pesi2}), the density of the data is a mixture of mixtures of Multinomial-Dirichlet distributions:

\begin{eqnarray}\label{DMoU.final}
 p(\x) =  \sum_{i=1}^{k_1} \sum_{j=1}^{k_2} \pi_{i|j}^{(1)} \pi_{j}^{(2)} \frac{\sum_{t=1}^{T} x_{t}}{\prod_{t=1}^{T} x_{t}} \frac{B\left(\sum_{t=1}^T x_t,\sum_{t=1}^T \beta_{it}(1+\alpha_{jt})\right)}{\prod_{t=1}^T B\left( x_t, \beta_{it}(1+\alpha_{jt})\right)}.
\end{eqnarray}

In other terms, adding a level to the hierarchy resulted in adding a mixing step. A double mixture is deeper, more flexible and it can capture more heterogeneity of the data, than a simple mixture of Multinomial-Dirichlet distributions. Having two layers and two number of groups for each, that are $k_1$ and $k_2$, it is important to define the procedure by which the units are clustered.

\subsection{Cluster assignment}
Theoretically, under this double mixtures, we could group units into $k_1$ groups, $k_2$ groups or $k_1 \times k_2$ groups. However, note that, under the constraint $-1 < \alpha_{jt} < 1$, the role of $k_2$ components at the deepest layer is confined to add flexibility to the model, while the real cluster-distribution is specified by the $\boldsymbol\beta$ parameters. For this reason, the number of `real' clusters is given by $k_1$. Their internal heterogeneity is captured by the $k_2$ sub-groups that help in adding more flexibility to the model. Therefore, the final allocation of the documents to the clusters is thus given by the posterior probability $p(z^{(1)}|\x) $ that can be obtained as follows:

\begin{eqnarray}\label{DMoU.cl}
 p(z^{(1)}=i|\x)=
  \frac{\sum_{j=1}^{k_2} \pi_{i|j}^{(1)} \pi_{j}^{(2)} p(\x|z^{(1)}=i,z^{(2)}=j)}{\sum_{i=1}^{k_1}\sum_{j=1}^{k_2} \pi_{i|j}^{(1)} \pi_{j}^{(2)} p(\x|z^{(1)}=i,z^{(2)}=j)}.
\end{eqnarray}

The model encompasses the simple MoU, that can be obtained as special case when $k_2=1$ and without any prior on $\boldsymbol \omega$. When $k_2=1$ and a Dirichlet prior is put on $\boldsymbol \omega$ a mixture of Dirichlet-Multinomial is defined. In this case, in order to assure identifiability, we assume $\boldsymbol\alpha=0$.

The approach can be generalized to multilayer of latent variables, where at each layer perturbation parameters to the final $\boldsymbol\beta$ are introduced, under the constraint that their values are limited between -1 and 1. However, we will show in the next Sections that the structure with just one latent layer is generally sufficient to gain large flexibility and very good clustering performance.

\section{Model Estimation}

 In this section we present a Bayesian algorithm for parameter estimation.
 The prior distribution for the weights of the mixture components are assumed to follow a Dirichlet distribution with hyperparameter $\delta$. We want non-informative priors for the model parameters. Hence, the value of the Dirichlet hyperparameter is $\delta=1$ in order to have a flat Dirichlet distribution. The prior distributions for each $\boldsymbol \alpha_j$ and $\boldsymbol \beta_i$ are given by the Uniform in the interval [-1,1] and in $(0,1000]$, respectively.

By using the previous model assumptions, the posterior distribution can be expressed as
 \begin{eqnarray*}
 \begin{aligned}
  p&(z^{(1)},z^{(2)},\boldsymbol\pi^{(1)},\boldsymbol\pi^{(2)},\boldsymbol\alpha,\boldsymbol\beta|\X) \propto  \\
  & p(\X|z^{(1)},z^{(2)},\boldsymbol\alpha,\boldsymbol\beta)
  p(z^{(1)}|z^{(2)},\boldsymbol\pi^{(1)})p(z^{(2)}|\boldsymbol\pi^{(2)})p(\boldsymbol\alpha)p(\boldsymbol\beta)p(\boldsymbol\pi^{(1)})p(\boldsymbol\pi^{(2)}),
\end{aligned}
\end{eqnarray*}

\noindent where $p(\X|z^{(1)},z^{(2)},\boldsymbol\alpha,\boldsymbol\beta)$ is the likelihood function of the model. By indexing the documents for which it holds that $z^{(1)}_{di}\cdot z^{(2)}_{dj}=1$, as $d: z^{(1)}_{di}\cdot z^{(2)}_{dj}=1$, the likelihood function can be expressed as:

 \begin{eqnarray}\label{DMoU.lik}
  p(\X|z^{(1)},z^{(2)},\boldsymbol\alpha,\boldsymbol\beta)=\prod_{i=1}^{k_1}\prod_{j=1}^{k_2}\prod_{d: z^{(1)}_{di}\cdot z^{(2)}_{dj}=1}\frac{\sum_{t=1}^{T} x_{dt}}{\prod_{t=1}^{T} x_{dt}} \psi_d(\boldsymbol\alpha_j,\boldsymbol\beta_i),
\end{eqnarray}
\noindent with $$\psi_d(\boldsymbol\alpha_j,\boldsymbol\beta_i)= \frac{B\left(\sum_{t=1}^T x_{dt},\sum_{t=1}^T \beta_{it}(1+\alpha_{jt})\right)}{\prod_{t=1}^T B\left( x_{dt}, \beta_{it}(1+\alpha_{jt})\right)}.$$

In order to sample parameters and latent variables from the posterior distribution we determine the full conditionals of each unobservable variable given the other ones.

\subsection{Full Conditionals}
The posterior distribution of the parameters and latent allocation variables given the other variables are proportional to known quantities. By using
$|\ldots$ to denote conditioning on all other variables, they are:

{\small
\begin{eqnarray*}\label{DMoU.fcond}
\hspace*{-0.5cm}
\begin{aligned}
 &p(\boldsymbol\pi^{(2)}|\ldots)\sim Dirichlet\left(\sum_{d=1}^n z_{d1}^{(2)}+\delta,\ldots,\sum_{d=1}^n z_{dj}^{(2)}+\delta,\ldots,\sum_{d=1}^n z_{dk_2}^{(2)}+\delta\right);\\
  &p(\boldsymbol\pi_{\cdot|j}^{(1)}|\ldots)\sim Dirichlet\left(\sum_{d=1}^n z_{d1}^{(1)}z_{dj}^{(2)}+\delta,\ldots,\sum_{d=1}^n z_{di}^{(1)}z_{dj}^{(2)}+\delta,\ldots,\sum_{d=1}^n z_{dk_1}^{(1)}z_{dj}^{(2)}+\delta\right);\\
  &Pr(z_{dj}^{(2)}=1|\ldots)\propto \prod_{i=1}^{k_1}\left(\pi_{i|j}^{(1)}\pi_{j}^{(2)}\psi_d(\boldsymbol\alpha_j,\boldsymbol\beta_i)\right)^{z_{di}^{(1)}};\\
  &Pr(z_{di}^{(1)}=1|\ldots)\propto \prod_{j=1}^{k_2}\left(\pi_{i|j}^{(1)}\psi_d(\boldsymbol\alpha_j,\boldsymbol\beta_i)\right)^{z_{dj}^{(2)}};\\
  &f(\boldsymbol\alpha_{j}|\ldots) \propto \prod_{i=1}^{k_1} \prod_{d: z^{(1)}_{di}\cdot z^{(2)}_{dj}=1}\psi_d(\boldsymbol\alpha_j,\boldsymbol\beta_i)\indicator{-1<\boldsymbol\alpha_{j}<1};\\
  &f(\boldsymbol\beta_{i}|\ldots) \propto \prod_{j=1}^{k_2} \prod_{d: z^{(1)}_{di}\cdot z^{(2)}_{dj}=1}\psi_d(\boldsymbol\alpha_j,\boldsymbol\beta_i)\indicator{0<\boldsymbol\beta_{i}<\infty}.
\end{aligned}
\end{eqnarray*}
}

A Gibbs sampling MCMC algorithm can be thus easily implemented for generating values from the posterior distributions.

Note that in order to get $\boldsymbol\alpha_j$ and $\boldsymbol\beta_i$ we need an accept-reject mechanism. We consider as proposal value for $\beta_i$, the average value of the parameters generated by $n \times k_2$ Dirichlet-Multinomials, given $\alpha_j$ fixed and viceversa.

The computational time of the algorithm depends only on the desired number of runs of the MCMC algorithm, the number of nodes $k_1$ and $k_2$ and on the length of the vectors $\boldsymbol\alpha_j$ and $\boldsymbol\beta_i$. For a two-class dataset, described in Section 6, with $D=240$, $T=357$ and $k_2=2$ the MCMC algorithm with 5.000 iterations requires about 10 minutes on
a processor Intel(R) Core(TM) i7-6500U CPU @ 2.50GHz, 2592 Mhz, 2 cores, under R cran 3.6.1.

We will show the estimation and clustering performance through a simulation study and real applications in the next Sections.

\section{Simulation Study}
The performance of the proposed method is evaluated under different aspects in an empirical simulation study. In order to prove the capability of the deep MoU to uncover the clusters in complex data, data were generated with a high level of sparsity. Several simulation studies are presented and discussed in the following.

The first simulation study aimed to check the capability of the deep MoU to cluster well the data, when these are generated according to a deep generative process. More precisely, we set $T=200$ and $n=200$, $k_1=3$, $k_2=2$ and balanced classes. We randomly generated $\boldsymbol\beta$ from a Uniform distribution in (0,20] and $\boldsymbol\alpha$ from a Uniform distribution in [-1,1]. In order to assure a high level of sparsity, for each document the total number of terms has been generated according to a Poisson distribution with parameter $N_d=20$, $\forall d$. Data are then organized in a document-term matrix containing the term frequencies of each pseudo-document.

\begin{figure}[t]
  \centering
  \includegraphics[width=1\textwidth]{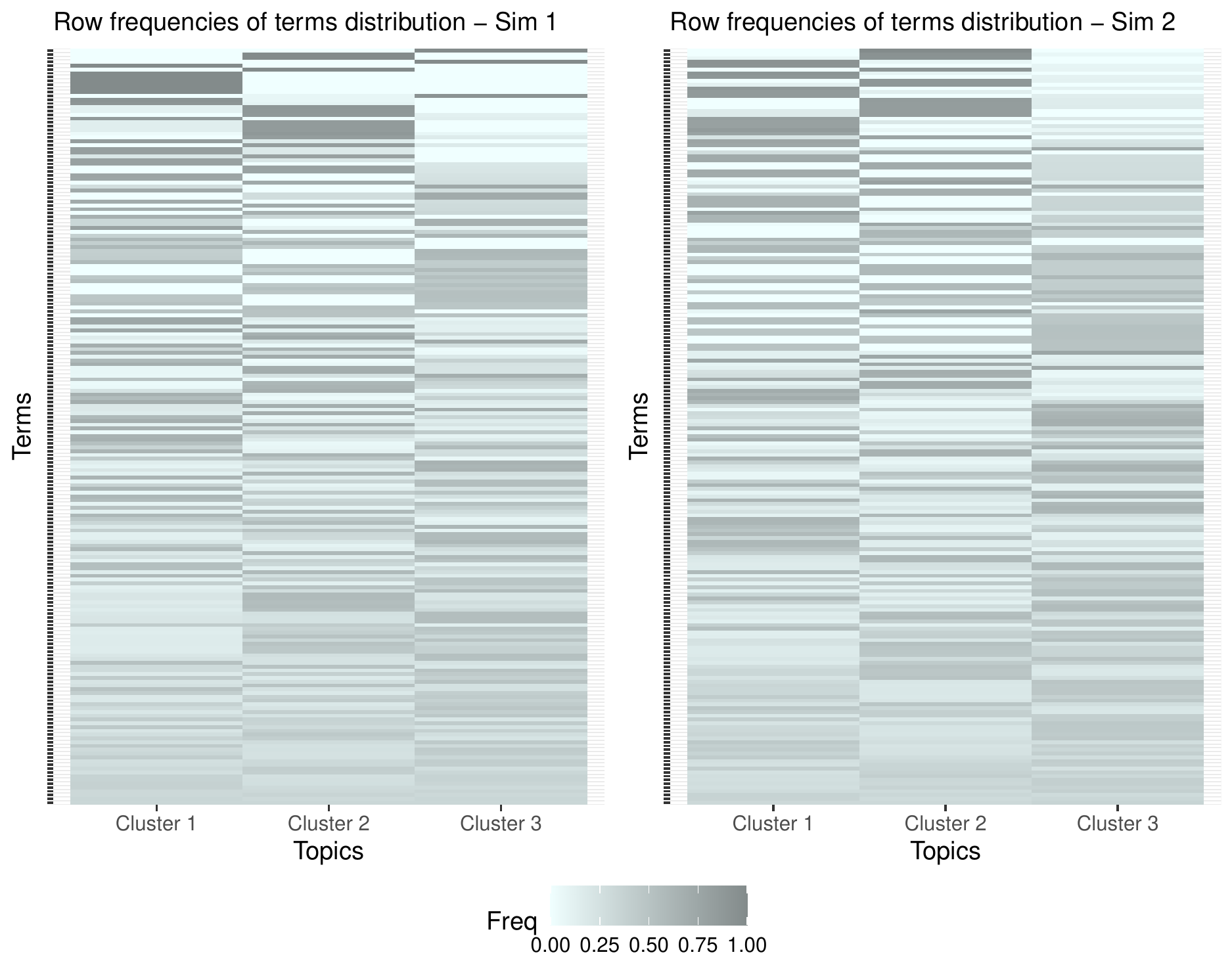}
  \caption{Heatmap of the conditional distribution of features across the classes. The left panel refers to the first simulation study and the right panel to the second ones.}\label{heatmap_sim}
\end{figure}

Panel (a) of Figure \ref{heatmap_sim} shows the row frequency distributions of the features across the clusters and provides a representation of the group overlapping.

Clustering performance has been measured by using both the Adjusted Rand Index (ARI) and the accuracy rate. The former is a corrected-for-chance version of the Rand Index \citep{hubert1985comparing}, that measures the degree to which two partition of objects agree; \cite{romano2016adjusting} proved that this measure is particularly indicated in presence of large equal sized reference clusters. The latter is defined as the ones'complement of the misclassification error rate.  Table \ref{tab.s1} shows the Adjusted Rand Index and the accuracy obtained by a deep MoU model for different values of $k_2$, ranging from 1 to 5. We run an MCMC chain with 5000 iterations, discarding the first 2000 as burn-in. Visual inspection assured that this burn-in was largely sufficient.

\begin{table}[ht]
\caption{Simulated Data, simulation study 1. Adjusted Rand Index (ARI) and Accuracy of Deep MoU for $k_1=3$ and different values of $k_2$.\label{tab.s1}}
\centering
\begin{tabular}{lcc}
  \hline
 $k_2$ & ARI & Accuracy \\
  \hline
1	& 0.468	&	0.755  \\
2	& 0.940	&	0.980  \\
3	& 0.910	&	0.970 \\
4	& 0.934	&	0.980 \\
5	& 0.925	&	0.975 \\
   \hline
\end{tabular}
\end{table}

\begin{table}[ht]
	\caption{Simulated Data, simulation study 2. Adjusted Rand Index (ARI) and Accuracy of Deep MoU for $k_1=3$ and different values of $k_2$.\label{tab.s2}}
	\centering
	\begin{tabular}{lcc}
		\hline
		$k_2$ & ARI & Accuracy \\
		\hline
		1	& 0.824	&	0.940  \\
		2	& 0.811	&	0.935  \\
		3	& 0.798	&	0.930 \\
		4	& 0.811	&	0.935 \\
		5	& 0.797	&	0.930 \\
		\hline
	\end{tabular}
\end{table}

The results show how the model with $k_2>1$ is really effective in clustering the data and, as desirable, the model with $k_2=2$ (i.e. the setting that reflects the generative process of the data) resulted to be the best one in terms of recovering the `true' grouping structure. The gap between $k_2=1$ and $k_2>1$ is relevant, however the performance remains elevate for the various $k_2>1$, thus indicating that a deep structure can be really effective in clustering such kind of data.

In a second simulation study we tested the performance of a deep MoU with data that are not originated according to a deep generative process, but simply by $k_1=3$ balanced groups, $T=n=200$ and $N_d=20$, $\forall d$.
Panel (b) of Figure \ref{heatmap_sim} displays the row frequency distributions of the generated features.
As shown in Table \ref{tab.s2}, in this situation, the deep model with $k_2>1$ does not significantly improve the clustering performance, and the accuracy remains stable as $k_2$ increases.
This suggests that when the data are pretty simple, and are not high-dimensional, a deep algorithm is not more efficient than the conventional MoU.

The third simulation study aimed at measuring the accuracy of the estimated parameters $\boldsymbol\alpha$ and $\boldsymbol\beta$ in data with double structure $k_1=3$ and $k_2=2$, allowing for different combinations of $T$, $n$ and $N$, so as to measure the effect of data dimensionality and level of sparsity on the goodness of fit. We considered a total of 8 different scenarios generated according to the combinations of $T=\{100, 200\}$, $n=\{100, 200\}$ and $N=\{10, 20\}$. Table \ref{tab.s3} contains the Euclidean distance between the true parameter vectors and the posterior means, normalized over $T$.

\begin{table}[ht]      \small
\caption{Simulated Data, simulation study 3. Precision of the estimated parameters for different values of $n$, $T$ and $N$.\label{tab.s3}}
\centering
\begin{tabular}{cccccccc}
  \hline
n & T & N & $\frac{ || \boldsymbol\beta_1 -\hat{\boldsymbol\beta_1}||}{T} $ & $\frac{ || \boldsymbol\beta_2 -\hat{\boldsymbol\beta_2}||}{T}$  & $\frac{ || \boldsymbol\beta_3 -\hat{\boldsymbol\beta_3}||}{T}$ & $\frac{ || \boldsymbol\alpha_1 -\hat{\boldsymbol\alpha_1}||}{T}$ & $\frac{ || \boldsymbol\alpha_2 -\hat{\boldsymbol\alpha_2}||}{T}$ \\
  \hline
100&100&10& 5.016 & 5.357 & 4.737 & 0.488 & 0.596   \\
100&200&10& 5.360 & 5.754 & 5.455 & 0.511 & 0.516  \\
100&100&20& 3.983 & 3.917 & 3.166 & 0.340 & 0.329  \\
100&200&20& 5.396 & 4.261 & 4.220 & 0.375 & 0.392    \\
200&100&10& 3.999 & 4.188 & 5.061 & 0.317 & 0.370  \\
200&200&10& 4.630 & 4.540 & 5.934 & 0.439 & 0.441 \\
200&100&20& 3.275 & 5.488 & 3.205 & 0.261 & 0.291  \\
200&200&20& 3.912 & 3.435	& 3.982 & 0.311 & 0.315  \\
   \hline
\end{tabular}
\end{table}

As expectable, for a given $T$ the goodness of fit improves as the number of documents increases. The level of sparsity has a relevant role as well: when $N$ increases the documents are more informative and the parameter estimates become more accurate.

\subsection{Application to real data}
The effectiveness of the proposed model is demonstrated by using four textual datasets, including the introduced ticket data. We compare the Deep MoU with conventional clustering strategies: $k$-means \citep{lloyd1982} with cosine distance ($k$-means) and with Euclidean distance on data transformed according to Semantic Analysis (LSA $k$-means), Partition Around Medoids \citep[PAM,][]{kaufman2009finding}, Mixture of Gaussians on semantic-based transformed data \citep[MoG,][]{Peel}, Hierarchical clustering according to Ward's method \citep[HC,][]{murtagh2014ward}  with cosine dissimilarity, Latent Dirichlet Allocation \citep[LDA,][]{blei2003latent}, Mixtures of Unigrams \citep[MoU,][]{Nigam} estimated via the EM algorithm \citep{Dempster1977}, spectral clustering \citep[SpeCl,][]{ng2002spectral} and affinity propagation clustering \citep[AffPr,][]{frey2007clustering} with normalized linear kernel.

\begin{figure}[t]
  \centering
  \includegraphics[width=0.9\textwidth]{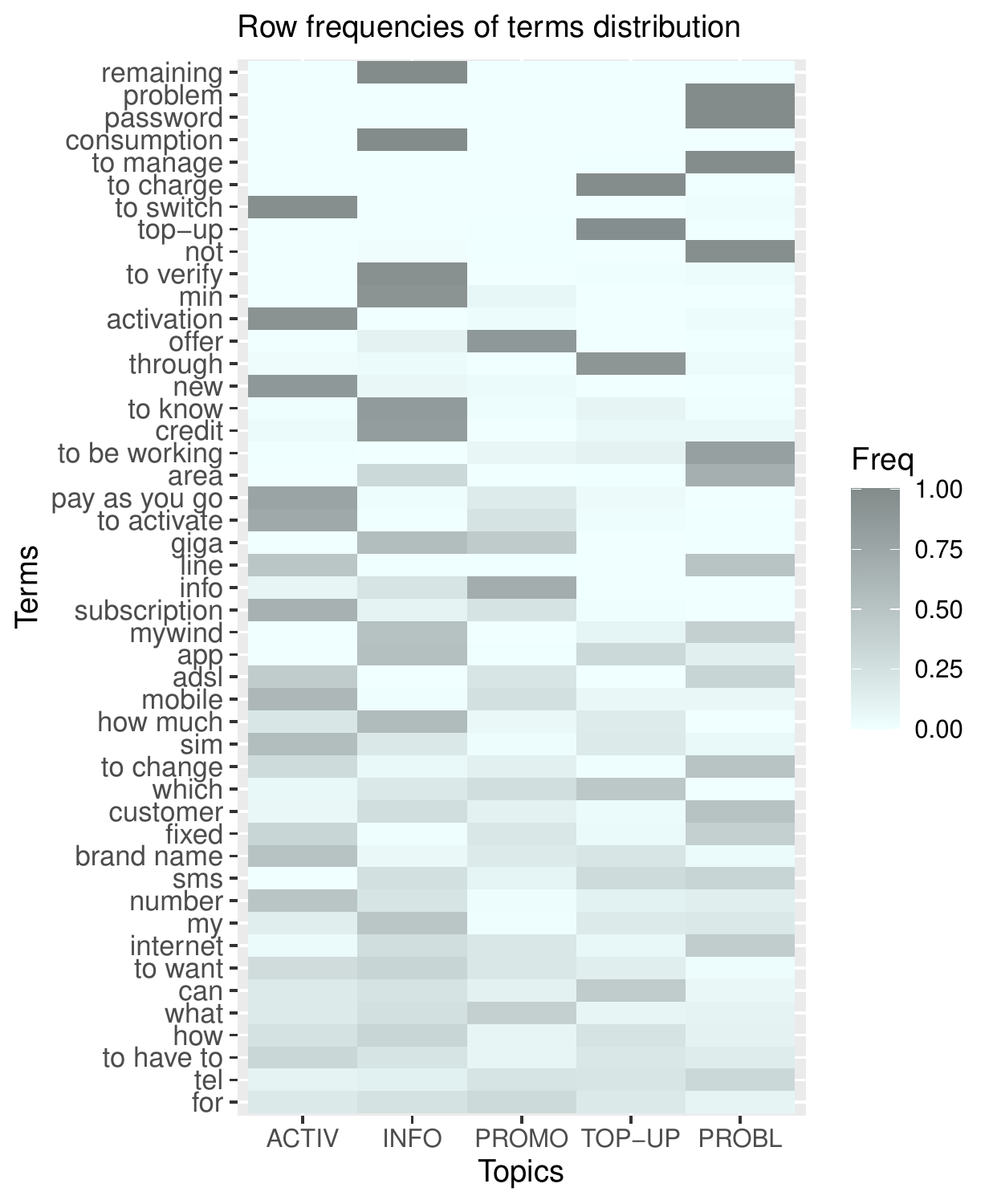}
  \caption{Heatmap of the conditional distribution of terms across classes. The shades reflect the distribution of row frequencies}\label{heatmap}
\end{figure}

The \verb"CNAE-9" dataset contains 1080 documents of free text business descriptions of Brazilian companies categorized into 9 balanced categories \citep{Ciarelli1,Ciarelli2} for a total of 856 pre-processed words. Since the classes 4 and 9 are the most overlapped we considered also the reduced \verb"CNAE-2" dataset composed by these two groups only which consists of 240 documents and 357 words. This data set is highly sparse: the 99.22\% of the document-term matrix entries are zeros.

The \verb"BBC" dataset consists of 737 documents from the BBC Sport website corresponding to sport news articles in five topics/areas (athletics, cricket, football, rugby, tennis) from 2004-2005 \citep{greene}. After a preprocessing phase aimed at discarding non-relevant words, the total number of features is 1075.
This dataset is moderately sparse with a fraction of zeros equal to 92.36\%.

The \verb"ticket" dataset contains $n=2129$ tickets and $T=489$ terms obtained after preprocessing: original raw data were processed via stemming, so as to reduce inflected or derived words to their unique word stem, and some terms have been filtered out in order to remove very common non-informative \emph{stopwords} words in the Italian language. The tickets have then been classified by independent operators to $k=5$ main classes described in Table \ref{tab.r1}.

\begin{table}[ht]                \small
\caption{Real Data. Number of tickets for each class.\label{tab.r1}}
\centering
\begin{tabular}{llc}
  \hline
 Class & Description & Freq. \\
  \hline
 ACTIV & Activation of SIM, ADSL, new contracts & 407 \\
  INFO & General information about current balance, consumption, etc. & 471 \\
  PROMO & Request of information about new offers and promotions& 376 \\
  TOP-UP & Top-up & 435 \\
  PROBL & Problems with password, top-up, internet connection, etc. & 440 \\
   \hline
\end{tabular}
\end{table}

\begin{figure}[b]
\centering
\hspace{-1.5cm}
  \includegraphics[width=1\textwidth]{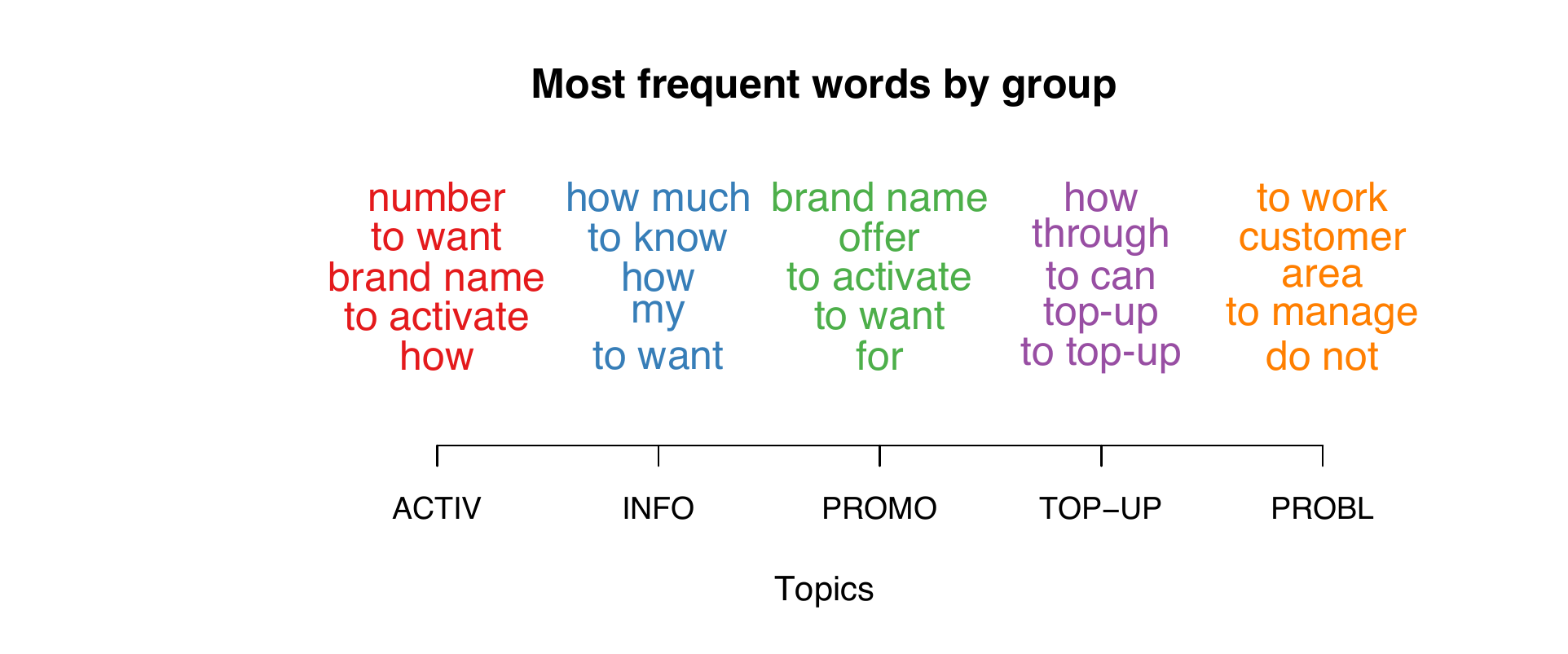}
  \caption{Most frequent terms within each group}\label{plottopic}
\end{figure}

The peculiarity and major challenge of this dataset is the limited number of words used, on average, for each ticket. In fact, after preprocessing, the tickets have an average length of 5 words. A graphical representation of the data, after the preprocessing step, is shown in Figure \ref{heatmap}. The heatmap shows the row frequency distributions of the most common terms (overall occurrences  $\geq$ 50). As clear from the shades, each class is characterized by a limited set of specific words, while the majority is uniformly distributed across them, thus making the classification problem particularly challenging.
In order to have further details about the terms characterizing the tickets within each topic, Figure \ref{plottopic} displays the most frequent words.

Data are characterized by a large amount of sparsity with the 99.05\% of zeros: as a consequence, most conventional clustering strategies fail.

Tables \ref{tab.acc} and \ref{tab.ari} show, respectively, the accuracy and the ARI of the different methods on the presented data sets. For comparative reasons, the true number of clusters was considered as known for all the methods.

\begin{table}[ht]     
\caption{Real Data. Accuracy (multiplied by 100) for different methods.\label{tab.acc}}
\centering
\begin{tabular}{lcccc}
  \hline
 Method & CNAE-2 & CNAE-9 & BBC & Ticket \\
  \hline
$k$-means & 55.8 & 59.8 & 76.5	&	51.6 \\
LSA $k$-means & 65.0 & 54.4 & 40.6	&	46.4 \\
PAM  & 52.5 & 11.9 &	28.8 &	22.2 \\
MoG	 & 50.4 & 37.6 &	43.6 &	43.8 \\
HC	 & 51.2 & 13.2 &	25.4 &	22.5 \\
LDA	 & 53.3 & 64.6 &	41.0 &	31.8 \\
MoU  & 60.8 & 69.6 &	59.0 &	74.3 \\
SpeCl& 72.9 & 50.3 &    44.5 &  53.0 \\
AffPr& 59.2 & 58.9 &    53.5 &  51.7 \\
   \hline
\end{tabular}
\end{table}

\begin{table}[ht]     
\caption{Real Data. Adjusted Rand Index (ARI) for different methods.\label{tab.ari}}
\centering
\begin{tabular}{lcccc}
  \hline
 Method & CNAE-2 & CNAE-9 & BBC & Ticket \\
  \hline
$k$-means     & \phantom{-}0.010 & \phantom{-}0.418 & \phantom{-}0.567 & \phantom{-}0.233\\
LSA $k$-means & \phantom{-}0.088 & \phantom{-}0.282 & \phantom{-}0.038 & \phantom{-}0.135\\
PAM  & \phantom{-}0.002 & \phantom{-}0.000  & \phantom{-}0.031 & \phantom{-}0.000\\
MoG	 & \phantom{-}0.000 & \phantom{-}0.092  & \phantom{-}0.060 & \phantom{-}0.158\\
HC	 & -0.003 & -0.005 & -0.004 & -0.001\\
LDA	 & \phantom{-}0.000 & \phantom{-}0.507 & \phantom{-}0.092 & \phantom{-}0.000\\
MoU  & \phantom{-}0.043 & \phantom{-}0.562 & \phantom{-}0.226 & \phantom{-}0.523\\
SpeCl& \phantom{-}0.207 & \phantom{-}0.284 & \phantom{-}0.104 & \phantom{-}0.224 \\
AffPr& \phantom{-}0.030 & \phantom{-}0.338 & \phantom{-}0.189 & \phantom{-}0.219 \\
   \hline
\end{tabular}
\end{table}

Among the considered methods, spectral clustering, $k$-means and mixtures of Unigrams are the most effective method for classifying the datasets. An advantage of MoU, which is inherited by the proposed Deep MoU, is that it seems to be not affected by the large number of zeros like the other methods, as it is based on proportions. The Latent Dirichlet Allocation model, despite its flexibility, is not able to improve the classification on these short documents, because it is based on the assumption of multiple-topics for each document, which is not realistic for short data, as in the case of ticket dataset.

We applied Deep MoU with $k_2=1,\ldots,5$. For each setting we run 5000 iterations of the MCMC algorithm, discarding the first 2000 as burn-in. From graphical visualization and diagnostic criteria we observed convergence and stability to the different choices of starting points and hyperparameters $\delta$, so we considered $\delta=1$.

Tables \ref{tab.acc2} and \ref{tab.ari2} contain the clustering results of the Deep MoU on the datasets, measured by accuracy and ARI respectively. The case $k_2=1$ corresponds to a Bayesian MoU and classification is improved with respect to the conventional MoU in most all empirical cases.
A probable reason for that is the adoption of Dirichlet-Multinomials (in Bayesian MoU) instead of classical Multinomials (in frequentist MoU), which are more flexible in modeling overdispersion with respect to the multinomial framework \citep{Wilson1991103}. In our analysis, the only exception is represented by the dataset CNAE-9, which is particularly challenging because it is composed by 9 unbalanced groups. In this case, classical MoU perform slightly better in terms of misclassification rate and ARI. This may depend on a particularly good starting point of the EM-algorithm for fitting MoU, which is initialized by the $k$-means clustering.

When we move from $k_2=1$ to a deeper structure with $k_2=2$ or $k_2=3$ the accuracy improves in all the analyzed experiments. This proves how the introduction of the parameters $\boldsymbol\alpha_j$ may be beneficial for classification. When the hidden nodes are greater than 3 ($k_2=4$ or $k_2=5$) results seem to be little worsened, probably because of the larger number of parameters to be estimated.

\begin{table}[ht]
\caption{Real Data. Accuracy (multiplied by 100) of Deep MoU for different number of nodes $k_2$.\label{tab.acc2}}
\centering
\begin{tabular}{lcccc}
  \hline
 $k_2$ & CNAE-2 & CNAE-9 & BBC & Ticket \\
  \hline
1	& 81.2 & 50.1 & 95.5 &	76.2 \\
2	& 93.3 & 73.3 &	95.8 &	89.2 \\
3	& 92.9 & 77.5 &	95.9 &	87.6 \\
4	& 92.5 & 77.3 &	93.9 &	85.6 \\
5	& 91.2 & 76.4 &	93.9 &	83.3 \\
   \hline
\end{tabular}
\end{table}

\begin{table}[ht]
\caption{Real Data. Adjusted Rand Index (ARI) of Deep MoU for different number of nodes $k_2$.\label{tab.ari2}}
\centering
\begin{tabular}{lcccc}
  \hline
 $k_2$ & CNAE-2 & CNAE-9 & BBC & Ticket \\
  \hline
1	& 0.388 & 0.290 & 0.888 &	0.616 \\
2	& 0.750 & 0.667 & 0.894 &	0.757 \\
3	& 0.736 & 0.661 & 0.895 &	0.721 \\
4	& 0.721 & 0.663 & 0.848 &	0.681 \\
5	& 0.679 & 0.615 & 0.843 &	0.651 \\
   \hline
\end{tabular}
\end{table}

\section{Final Remarks}
In this paper we have proposed a deep learning strategy that extends the Mixtures of Unigrams model. With respect to other clustering methods, MoUs have desirable features for textual data. Firstly, document-term matrices usually contain the frequencies of the words in each document; for this reason, MoUs represent an intuitive choice, since they are based on multinomial distributions that are the probabilistic distributions for modeling
positive frequencies. Moreover, as MoUs naturally model proportions, they are not affected by the large amount of zeros of the datasets like other methods, so they are a proper choice for modeling very short texts and sparse datasets. Furthermore, MoU are based on the idea that documents related to the same topic have similar distributions of terms, which is realistic in practice. Taking a mixture of $k$ multinomials means doing clustering into $k$ topics/groups: there is a unique association between documents and topics.

All these nice characteristic are inherited by the proposed Deep MoU model. The proposed Deep MoU is particularly effective in clustering with challenging issues (sparsity, short document length and high-dimensionality). Being hierarchical in its nature, the model can be easily estimated by a MCMC algorithm in a Bayesian framework. In our analysis we chose non-informative priors, because there is little prior information available on the empirical context. The estimation algorithm produces good results in all the simulated and real situations considered here.

The proposed model could be extended in several directions: as discussed in Section 3, several hidden layers (instead of just a single one) could be considered. The merging function $\boldsymbol\beta_i(1+\boldsymbol\alpha_j)$ has been defined for identifiability reasons under the idea that the number of estimated groups is $k_1$ and the latent layer is only aimed at perturbing the $\boldsymbol\beta_i$ parameters for capturing some residual heterogeneity inside the groups. Of course, more complex nonlinear functions could be considered, without however losing sight of identifiability. In case of non-extreme sparsity and long documents, the model could be also extended to allow for deep \emph{m}-gram models. We leave all these ideas to future research.

\bibliographystyle{chicago}
\bibliography{ref}%

\end{document}